\documentclass[letterpaper, 10 pt, conference]{ieeeconf}  % Comment this line out if you need a4paper

\IEEEoverridecommandlockouts                              % This command is only needed if 
\usepackage{graphics} % for pdf, bitmapped graphics files
\usepackage{epsfig} % for postscript graphics files
\usepackage{amsmath} % assumes amsmath package installed
\usepackage{amssymb}  % assumes amsmath package installed
\usepackage{gensymb}
\usepackage{bm}
\usepackage{eqnarray}
\usepackage{algorithm} 
\usepackage{algorithmic}  
\usepackage[algo2e]{algorithm2e} 
\usepackage{booktabs} % for professional tables
\usepackage{xcolor}
\newcommand{\norm}[1]{\left\lVert#1\right\rVert}

\title{\LARGE \bf
SimGAN: Hybrid Simulator Identification for Domain Adaptation via Adversarial Reinforcement Learning
}

\author{Yifeng Jiang$^{1,2}$, Tingnan Zhang$^{1}$, Daniel Ho$^{3}$, Yunfei Bai$^{3}$, C. Karen Liu$^{2}$, Sergey Levine$^{1,4}$ and Jie Tan$^{1}$% <-this % stops a space

\thanks{$^{1}$Robotics at Google, Mountain View, CA, USA}%
\thanks{$^{2}$Computer Science Department, Stanford University, Stanford, CA, USA}%
\thanks{$^{3}$Everyday Robots, X The Moonshot Factory, Mountain View, CA, USA}%
\thanks{$^{4}$EECS, University of California, Berkeley, Berkeley, CA, USA}%
\thanks{Code available at: https://github.com/jyf588/SimGAN Accompanying video: https://youtu.be/McKOGllO7nc}%

}

\usepackage{etoolbox}
\makeatletter
\patchcmd{\@makecaption}
  {\scshape}
  {}
  {}
  {}
\makeatother

\begin{document}

\maketitle
\thispagestyle{empty}
\pagestyle{empty}

%%%%%%%%%%%%%%%%%%%%%%%%%%%%%%%%%%%%%%%%%%%%%%%%%%%%%%%%%%%%%%%%%%%%%%%%%%%%%%%%
\begin{abstract}
As learning-based approaches progress towards automating robot controllers design, transferring learned policies to new domains with different dynamics (e.g. sim-to-real transfer) still demands manual effort. This paper introduces SimGAN, a framework to tackle domain adaptation by identifying a hybrid physics simulator to match the simulated trajectories to the ones from the target domain, using a learned discriminative loss to address the limitations associated with manual loss design. Our hybrid simulator combines neural networks and traditional physics simulation to balance expressiveness and generalizability, and alleviates the need for a carefully selected parameter set in System ID. Once the hybrid simulator is identified via adversarial reinforcement learning, it can be used to refine policies for the target domain, without the need to interleave data collection and policy refinement. We show that our approach outperforms multiple strong baselines on six robotic locomotion tasks for domain adaptation.

\end{abstract}

%%%%%%%%%%%%%%%%%%%%%%%%%%%%%%%%%%%%%%%%%%%%%%%%%%%%%%%%%%%%%%%%%%%%%%%%%%%%%%%%
\section{Introduction}

%%yifeng: three discuss items for intro

Using simulation data to train a reinforcement learning (RL) policy for robotic skills provides an effective way to acquire diverse behaviors \cite{akkaya2019solving, RoboImitationPeng20}. While learning-based approaches have made controller design in simulation more automatic and less demanding of domain-specific knowledge, transferring a trained policy from simulation to the real hardware is often a heavily manual process. For example, with domain randomization techniques, practitioners select aspects of the simulation to randomize and train policies that are robust across a wide range of dynamics. Such ranges should be large enough to cover the unmodeled discrepancies between the simulation and real domains, while not being overly large as to adversely impact task performance. 

Alternatively, system identification methods (System ID) have the promise to improve performance by utilizing observations from the real hardware to fit model parameters accurately. However, they make the strong assumption that the system is parameterized by a set of predefined parameters, and the true model lies within the model class. Further, when designing a similarity loss between simulated and real trajectories, the commonly used $l_p$-norm assumes an additive error model which can be unrealistic, since errors from misidentified dynamics may not be stochastic but rather systematic. As such, heuristically crafting such a metric between paired trajectories could require additional engineering effort.

In this paper, we take a different approach and ask: can we identify the simulator to the extent that simulated trajectories are hard to distinguish from real ones, without manual design of randomization parameters or heuristic assumptions about model classes or model noise? We propose a new method for simulation identification, in which a Generative Adversarial Network (GAN) \cite{goodfellow2014generative} discriminator distinguishes between training and target domains and provides a learned similarity loss. In addition to reducing manual effort for loss design, a learned discriminative loss also lifts the requirement of calculating loss on \emph{paired} trajectories. Instead, the GAN loss incentivizes trajectory matching on the distribution (set of trajectory tuples) level \cite{zhu2020unpaired}. This allows system identification with excitation trajectories of \emph{variable} lengths which could be unstable or sensitive to initial conditions.

The adversarial learning framework provides us an objective for system identification, but we still need to select a proper parameterization. Simply learning a set of constant parameters in a standard simulator may not be sufficiently expressive to capture unmodeled phenomena. On the other hand, discarding all physics knowledge and learning a black-box model may cause model degradation when the state-action distribution shifts during policy refinement. We instead construct a hybrid simulator, where we replace the constant simulation parameters with a differentiable state-action-dependent function. Since we want to minimize the GAN loss over each of the trajectory tuples, similar to maximizing accumulated rewards in reinforce learning, we treat our simulation parameter function as an ``RL agent'' and optimize it using policy gradient methods. With such parameterization, we are able to only collect data once for the identification without the need to interleave it with policy training.

In summary, our contributions include: 
\begin{itemize}
    \item A novel formulation of simulation identification as an adversarial RL problem;
    \item A learned GAN loss that provides weak set-level supervision to alleviate issues associated with manual loss design and sensitive excitation trajectories;
    \item An expressive hybrid simulator parameterization to alleviate the need for a carefully selected parameter set.
\end{itemize}  
Our method can be used for domain adaptation of RL policies, by first identifying the simulator to match target domain trajectories and then refining the suboptimal behavior (data-collection) policy under the learned simulator. We evaluate our method against multiple strong baseline methods on two simulated robots and three target environments with different dynamics from the source environment\footnote{Due to COVID-19, physical robots are not accessible.}. We also show that the same simulator learned with our method can be used for training multiple different motor skills.

\section{Related Work}

To overcome the dynamics discrepancies which renders policies trained in simulations to fail in the real world, domain randomization techniques learn policies from simulation with randomized physical parameters. This results in robust policies that could successfully bridge the sim-to-real gap \cite{Sim2Real2018, pinto2017robust, andrychowicz2020learning, tan2018simtoreal}.
%%SL.10.30: this seems to be missing the foundational papers on this? sadeghi, tobin, etc. (should be from 2016/2017) it might be nice for a robotics audience to also include citations to robust control work, which in some sense did basically the same thing but much earlier, and is arguably more relevant, since both sadeghi and tobin were concerned with visuals, not physics (and tobin didn't even do randomization for control at all)
% add tobin and sadeghi to visual dr paragraph.
As overly wide randomization may hurt task performance, algorithms have been proposed to automatically adapt the sampling distribution of parameters throughout the course of training. For example, \cite{akkaya2019solving} used task performance as a metric to design a curriculum for gradually expanding randomization ranges. \cite{mehta2019active} used GAN to distinguish difficult dynamics from easier ones, and biased the sampling towards difficult parameters. Although we also use GAN in this paper, it is for a different purpose of distinguishing dynamics between the source and the target domains.

Although domain randomization does not need any target domain data, collecting a small amount of data in the target domain can significantly improve the performance of adapted policies. These methods include fine-tuning \cite{pmlr-v78-rusu17a}, sample-efficient model-based learning \cite{yang2020data, IROS20-Desai, song2020provably}, meta-learning \cite{song2020rapidly, nagabandi2018learning}, and latent space methods \cite{yu2019simtoreal, yu2020learning, RoboImitationPeng20}. These few-shots adaptation methods often interleave data collection with policy improvement to maximize data efficiency and avoid distribution shift. Similar to our method, \cite{eysenbach2020offdynamics} trained a domain classifier to tackle adaptation, focusing on scenarios when the target domain is more constrained than source. While we also assume limited target domain access in this work, we focus on whether our learned dynamics can generalize with only one batch of data collection, to further reduce manual burden with data collection. 

System identification is an effective way to improve simulation fidelity, and thus narrows the sim-to-real gap \cite{AAMAS13-Farchy, tan2016simulation}. Researchers have examined the identifiability of each robot parameter and designed maximally informative excitation trajectories to reveal these parameters \cite{doi:10.1177/0278364913495932, doi:10.1177/027836499201100408, jegorova2020adversarial}. Recent works \cite{rajeswaran2017epopt, chebotar2019closing, muratore2020bayesian, zhu2018fast, desai2020imitation} focused on if the identified model can improve policy performance, rather than seeking the ground-truth physical parameters. \cite{chebotar2019closing, muratore2020bayesian} identified distributions of simulator parameters, \cite{9341701} learned a state-dependent similarity loss by formulating a meta-learning problem using multi-task training data, and \cite{jeong2019modelling} identified generalized forces using RL as the optimizer. Although our work also uses RL, we adopt a GAN loss to better handle possibly large mismatches between paired trajectories than the mean-squared-errors, and use a different simulator parameterization which is in between of the previous works.

While our work aims at bridging the domain gap on dynamics, the same challenge also appears when transferring policies with visual input \cite{sadeghi2016cad2rl, tobin2017domain, ruiz2019learning}. GAN-based techniques have shown promises in solving the problem of domain adaptation for visual sim-to-real gaps \cite{bousmalis2017using,james2019simtoreal,rao2020rlcyclegan}.

\section{Simulator Identification via Adversarial RL}

%%yifeng: discuss point for method:
% 1. the order of subsections.
% how to handle the equations

\begin{figure*}[t]
  \centering
  \includegraphics[scale=0.3]{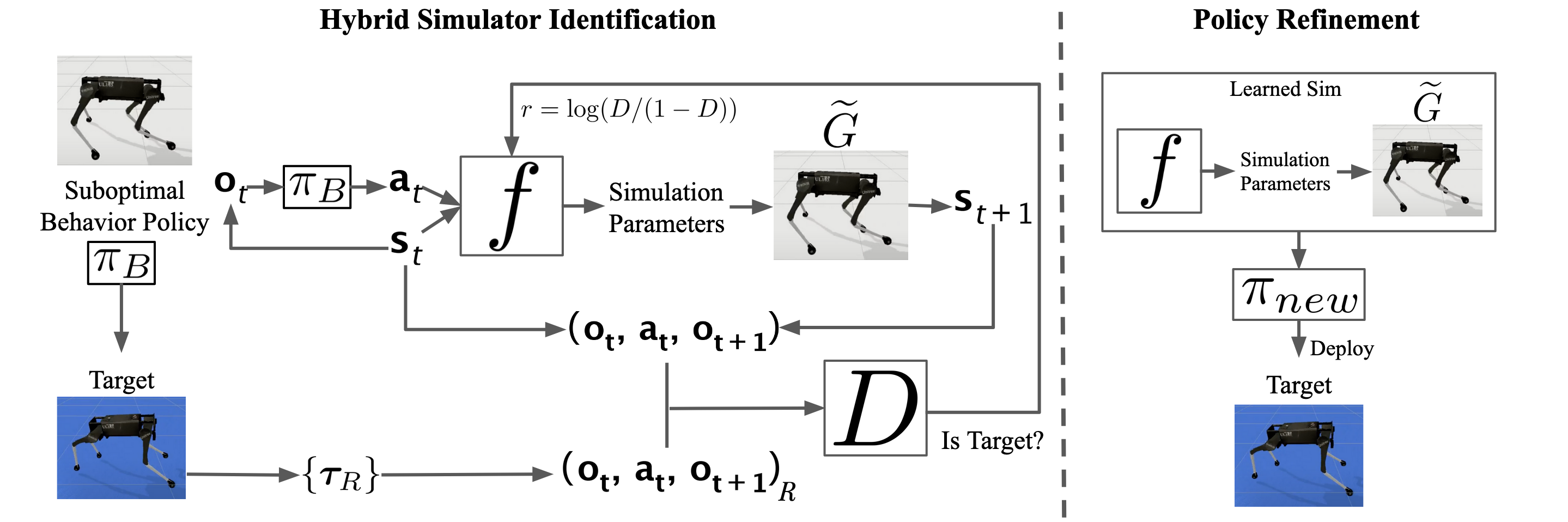}
  \caption{\label{fig:simGan} Algorithm Overview: (Left) We propose to learn a hybrid simulator using Adversarial RL, with $f$ being a learnable state-action-dependent simulation parameter function and $\widetilde{G}$ being an analytical simulator. Specifically, we collect a small amount of target domain data $\{ \tau_R\}$ using a behavior policy $\pi_B$. A discriminator $D$ is co-trained with $f$ using standard GAN framework. (Right) The learned hybrid simulator is used as the training environment to train a new policy $\pi_{new}$ that can achieve superior performance in the target domain. No target domain data are needed in this stage.}
\end{figure*}

Our method (Fig. \ref{fig:simGan}) optimizes a hybrid simulator via reinforcement learning to match its generated trajectories to those recorded in the target environment, with the similarity reward provided by a co-learned GAN discriminator. A hybrid simulator, compared with a parameterized analytical simulator or a purely statistical dynamics model, strikes a balance between model expressiveness and maximizing the state-action space where the model is physically valid. By using a GAN-style loss rather than the $l_p$ norm, we eliminate the need to align trajectories by abstracting the loss between paired trajectories to one between distributions. 

%%SL.10.30: You could borrow some phrasing from the paragraph I edited in the intro to talk about additive errors
% An additive noise error model can mislead the reinforcement learning process, since it assigns a non-zero probability to transitions that may in fact be impossible, such as injecting energy into a rigid body system or violating complementarity conditions for contacts.
%% the infeasible physics part does not really apply to us here since we are learning physics parameters.

\subsection{Background}

\textbf{Trajectory Representation:} A trajectory is a sequence $\bm{\tau} = (\bm{o}_0, \bm{a}_0, … \bm{o}_T)$, where $\bm{o}_t \in \mathcal{O}, \bm{a}_t \in \mathcal{A}$ are observations and actions respectively. With the initial state $\bm{s}_0 \sim p_0(\bm{s}_0)$ sampled from an initial state distribution in a state space $\mathcal{S}$, an observation function $g(\cdot): \mathcal{S} \rightarrow \mathcal{O}$, a dynamics model $\bm{s}_{t+1} \sim P(\bm{s}_{t+1} | \bm{s}_t, \bm{a}_t)$, and a policy given by $\bm{a}_t \sim \pi(\bm{a}_t | \bm{o}_t)$, the probability of generating one trajectory $p(\bm{\tau})$ is given by $p_\pi(\bm{\tau}) = p_0(\bm{s}_0)g(\bm{o}_0|\bm{s}_0)\prod \limits_{t=0}^{T-1} g(\bm{o}_{t+1}|\bm{s}_{t+1})P(\bm{s}_{t+1}|\bm{s}_t, \bm{a}_t)\pi(\bm{a}_t|\bm{o}_t).$
% A trajectory can come from the source (unmodified simulation), training (hybrid simulation), or target (real) domain. While full state $\bm{s}_t$ are available in simulations, this assumption is unrealistic in the target domain. For example, on a real robot, we can only observe a subset of the state from onboard sensors. An observation function $g(\cdot): \mathcal{S} \rightarrow \mathcal{O}$ maps states to partially observable observations $\bm{o}_t \in \mathcal{O}$ and policies takes $\bm{o}_t$ as input -- $\bm{a}_t \sim \pi(\bm{a}_t | \bm{o}_t)$, leading to an \textit{observed} trajectory denoted as ${\bm{\tau}} = (\bm{o}_0, \bm{a}_0, … \bm{o}_t)$.

\textbf{Hybrid Simulator:} A hybrid simulator combines an analytical simulator and learnable components. Compared with a parameterized analytical simulator $\bm{s}_{t+1} = G(\bm{s}_t, \bm{a}_t; \bm{c})$ where $\bm{c}$ are the simulation parameters, in a hybrid simulator that is represented as $\bm{s}_{t+1} \sim \widetilde{G}(\bm{s}_{t+1} | \bm{s}_t, \bm{a}_t, \bm{c}_t)$, $\bm{c}_t$ can be learned -- for instance $\bm{c}_t \sim f_{\bm{\theta}}(\bm{c}_t | \bm{s}_t, \bm{a}_t)$ where $f_{\bm{\theta}}$ is a learnable function, such as a neural network with weights $\bm{\theta}$. Note that $\bm{c}_t$ is general, and can represent or replace any component(s) in the analytical simulator. For example, $\bm{c}_t$ can represent actual motor forces or contact forces, modifying or replacing those solved by the simulator. Trajectory probabilities under a hybrid simulator is given by
\begin{align}
\label{eq:traj_prob}
p_{\pi, f}(\bm{\tau}) =& p_0(\bm{s}_0)g(\bm{o}_0|\bm{s}_0)\left(\prod \limits_{t=0}^{T-1} \pi(\bm{a}_t|\bm{o}_{t})g(\bm{o}_{t+1}|\bm{s}_{t+1}) \right.\nonumber \\
             & \left.\widetilde{G}(\bm{s}_{t+1}|\bm{s}_t, \bm{a}_t, \bm{c}_t)f_{\bm{\theta}}(\bm{c}_t | \bm{s}_t, \bm{a}_t)\right).
\end{align}

\subsection{Hybrid Simulator Identification}

The goal of hybrid simulator identification is to learn the function $f$ to maximize the similarity of distribution between the transition tuples that are generated from our hybrid simulator and those collected from target domains. In this work, we first use a fixed, suboptimal \textit{behavior policy} $\pi_B$ to collect $N$ offline trajectories $\{{\bm{\tau}_R}\}_{1,\cdots N}$ in the target environment, and then optimize for an $f_{\bm{\theta}}$ to generate similar simulated trajectories under the same $\pi_B$:

% The following texts are too convoluted to understand for general audience. Remove it for now. If needed will add back by rephrasing.
% $f$ cannot be trained independently as we usually will not have access to target domain $\bm{c}_t$ values. That said, observing the formula for $p(\bm{\tau})$, both $f$ and $\pi$ are differentiable, while $\widetilde{G}$ is usually not because of constraint solving algorithms. As such, we exchange the roles of $f$ and $\pi$, treating $f$ as an ``RL agent'' and $\widetilde{G}$ and $\pi$ as the ``environment'', so that we can optimize $f$ with policy gradient methods.

% The learned component $f$, though being differentiable, cannot be trained independently as we usually will not have access to target domain $\bm{c}_t$ values. Instead we must optimize $f$ outside the simulator $\widetilde{G}$ with its generated observations and actions. As $\widetilde{G}$ is oftentimes non-differentiable because of constraint solving algorithms, we treat $f$ as an "RL agent" and $\widetilde{G}$ and $\pi$ as the "environment" and optimize $f$ with policy gradient methods. In other words, we exchanged the roles of $f$ and $\pi$ in the formula of $p(\bm{\tau})$.

% Our method will not introduce theoretical difficulty to existing adversarial RL frameworks, since it should be clear from Eq. \ref{eq:prob-traj} that the roles of “transition dynamics” and “policy” can be exchanged.
% The effects of different $f$ can only be observed from outside of the simulator, which can have non-differentiable components but can be optimized with RL. 

\begin{eqnarray}
\label{eq:prob-traj}
\max_{{\bm{\theta}}} J(f_{\bm{\theta}}) &=& \mathbb{E}_{\bm{\tau}} \sum_{t=0}^T \gamma^t r(\bm{o}_t, \bm{a}_t, \bm{o}_{t+1}) \\
\textrm{s.t.}& &\bm{\tau} \sim p_{\pi_B, f}(\bm{\tau}) \textrm{  in Eq. (\ref{eq:traj_prob}}), \nonumber
\end{eqnarray}
% \begin{eqnarray*}
% \max_{{\bm{\theta}}} J(f_{\bm{\theta}}) = \mathbb{E}_{{\bm{\tau}}} \sum_{t=0}^T \gamma^t r(\bm{o}_t, \bm{a}_t, \bm{o}_{t+1}) \\
% \textrm{s.t.~~} \bm{\tau} \sim p_0(\bm{s}_0) \prod \limits_{t=0}^T \pi_B(\bm{a}_t|\bm{o}_t)\widetilde{G}(\bm{s}_{t+1}|\bm{s}_t, \bm{a}_t, \bm{c}_t)f_{\bm{\theta}}(\bm{c}_t | \bm{s}_t, \bm{a}_t), \label{eq:prob-traj} \\
% \bm{o}_t \sim g(\bm{s}_t),
% \end{eqnarray*}
where $\gamma$ is the reward discount factor and the reward function $r(\bm{o}_t, \bm{a}_t, \bm{o}_{t+1})$ measures the similarities between tuples generated by $f$ and the target domain tuples from $\{{\bm{\tau}_R}\}$. Note that the optimization Eq. (\ref{eq:prob-traj}) resembles a typical formulation of reinforcement learning, with the only difference that the simulator function $f$ is now to be optimized while the policy $\pi_B$ is fixed. We purposefully use RL here because we would like to minimize the multi-step accumulated discrepancy between domains. To calculate the reward, we propose to adversarially train a discriminator $D_{\bm{w}}(d_t | \bm{o}_t, \bm{a}_t, \bm{o}_{t+1})$, parameterized by $\bm{w}$, to predict whether a tuple is generated from simulation (score $d=0$) or sampled from the target domain ${{\bm{\tau}_R}}$ (score $d=1$). Same as in previous works that combine GAN and RL \cite{fu2017learning, ho2016generative}, our reward is constructed from the current discriminator score $r_t= \log (d_t / (1-d_t))$, favoring $d$ values closer to 1. The training of the discriminator $D_{\bm{w}}$ alternates with the training of the generator $f$, as in standard GAN frameworks. 
% \todo{You are using D for discriminator score here, but using p in the Algorithm 1.} 

% The idea of using GANs in RL to solve an imitation learning problem is first proposed in GAIL \cite{ho2016generative}, and in fact our method can be most easily understood as “GAIL for dynamics”,
% %%SL.10.30: I don't particularly like this term, because we are *not* doing imitation learning. So I don't think our method is "GAIL for dynamics" but rather "GANs for training a simulator" -- that is, if GAIL is an application of GANs for imitation, ours is an application of GANs for simulator parameter identification. That is, they are adjacent, not that one follows from the other.
% where we fix policy $\pi_B$ and optimize the (partial) dynamics model $f$ to mimic recorded trajectories, rather than the opposite as in GAIL.

Depending on the sign of the reward, the learned dynamics can sometimes produce overly long or short trajectories in order to collect more rewards or avoid penalties, by sacrificing its accuracy. We propose a simple technique to offset $r$ with a positive alive bonus discouraging over-early termination when the generated trajectories are unrealistically short, and vice versa. Inspired by \cite{kostrikov2018discriminator}, we design this \textit{adaptive alive bonus} to be $b_i = \log(l_i/l_R)$, where $l_i$ is the average length of the simulated trajectories under current $f_i$, and $l_R$ is the average length of the target domain trajectories $\{{\bm{\tau}_R}\}$.

Finally, once the simulator component $f$ is learned, a new control policy $\pi_{new}$ is refined from $\pi_B$ using standard deep reinforcement learning in the hybrid simulator $\widetilde{G}$ with the learned $f$. All the data is generated using the learned hybrid simulator. The fine-tuned policy $\pi_{new}$ demonstrates strong performance in the target domain. Algorithm \ref{algo:general_zo_sgd} summarizes the key steps of our hybrid simulator identification approach.

\begin{algorithm}
\SetAlgoLined

\SetKwInput{KwInput}{Require}
\KwInput{Suboptimal Behavior Policy: $\pi_B(\bm{a_t}|\bm{o_t})$, \\ Physics Simulator: $\widetilde{G}(\bm{s}_{t+1}|\bm{s}_t, \bm{a}_t, \bm{c_t})$, \\
Initial Simulator Parameter Function: $f_{\bm{\theta_0}}(\bm{c_t}|\bm{s_t},\bm{a_t})$,  \\ 
Initial Discriminator: $D_{\bm{w_0}}(p|\bm{o}_t, \bm{a}_t, \bm{o}_{t+1})$ }
%\texttt{\\}
\vspace{5pt}
// Data collection in target domain. \;

Collect $N$ trajectories $\{\bm{\tau}_R\}_{1\cdots N}$ in Target by deploying $\pi_B$ \;

$l_R := \textrm{average-length}(\{\bm{\tau}_R\})$
\vspace{5pt}

// Hybrid simulator identification. \;

\For{i = 0,1,... n}{
    Sample trajectories $\{\bm{\tau}_{i}\}$ from $\pi_B$ under $\widetilde{G}$ and $f_{{\bm{\theta}}_i}$ \;
    
    Update $D$ from $\bm{w}_i$ to $\bm{w}_{i+1}$ with cross entropy loss:
    \begin{eqnarray*}
    % (\bm{o}_t, \bm{a}_t, \bm{o}_{t+1})
            -\hat{\mathbb{E}}_{\bm{\tau}_R}[\log D_{\bm{w}}] - \hat{\mathbb{E}}_{\bm{\tau}_i}[\log(1- D_{\bm{w}})]
    \end{eqnarray*}

    $l_i := \textrm{average-length}(\{\bm{\tau}_i\})$ \;
    
    Calculate adaptive alive bonus: $b_i := \log(l_i / l_R)$ \;
    
    Update $f$ from ${\bm{\theta}}_i$ to ${\bm{\theta}}_{i+1}$ using PPO, with reward $r_t$:
    $$
        d_t = D_{\bm{w}_{i+1}}(\bm{o}_t, \bm{a}_t, \bm{o}_{t+1})
    $$
    $$
        r_t = \log(d_t/(1-d_t)) + b_i
    $$
}

\vspace{5pt}
// Refining the policy in identified simulator. \;

$\pi_{new} := \pi_B$ \;

Train $\pi_{new}$ using PPO under $\widetilde{G}$ with $f_{{\bm{\theta}}_n}$  \;

\Return $\pi_{new}$
\caption{Simulator Identification and Policy Adaptation via Adversarial RL}
\label{algo:general_zo_sgd}
\end{algorithm}

\subsection{Parameterization of Hybrid Simulator}
\label{sec:parameterization}
%%SL.10.30: Check out some of the phrasing I suggested in the intro to discuss this.

%Simply learning the  constant parameters in a standard simulator may be insufficiently expressive to capture unmodeled phenomena, while discarding all physical knowledge and learning a black-box model may cause model degradation when the  state-action distribution shifts during policy adaptation. 

The parameterization of $\bm{c}_t \sim f$ controls how much of the simulator is learned and how much is inherited from the analytical simulator. On one hand, during policy adaptation, the state-action distribution shifts away from where $f$ is trained on, so $f$ needs to avoid model degradation and remain physically valid for a reasonably large region of state-action space. On the other hand, $\bm{c}_t$ should be expressive enough to capture a large range of unmodeled domain discrepancies. 

In this work, we model $\bm{c}_t \sim N(\bm{\mu}_t | (\bm{s}_t, \bm{a}_t), \bm{\sigma}_t | (\bm{s}_t, \bm{a}_t)))$ as a \textit{state-action-dependent, stochastic} simulation parameter function, where $(\bm{\mu}_t , \bm{\sigma}_t) = f_{\bm{\theta}}(\bm{s}_t, \bm{a}_t)$, meaning $f$ outputs the means and variances of the simulation parameters. Specifically, $\bm{c}$ include contact-related simulator parameters (friction, restitution, spinning friction, contact error reduction rate (ERP) of each body in contact) and scaling factors of each motor action (i.e., $\bm{\tau} = \bm{c}_a \bm{a}$, where $\bm{a}$ is torque output from the policy, $\bm{c}_a$ is the subvector of $\bm{c}$ representing the scaling factors, and $\bm{\tau}$ is the actual motor input passed to the analytical simulator). This choice reflects the fact that modeling contact \cite{yu2016more} and joint actuation \cite{hwangbo2019learning} are important for robotics tasks. But more importantly, state-action-dependent contact and actuator parameters respectively cover the spaces of external and internal forces in the equation of motions:
\begin{equation*}
    \Ddot{\bm{q}} = \bm{M(\bm{q})}^{-1} ( \bm{C}(\bm{q}, \dot{\bm{q}}) + \bm{\bm{\tau}} + \bm{J}^T(\bm{q}) \bm{p} ),
\end{equation*}
where $\bm{\bm{\tau}}$ and $\bm{p}$ are internal and external forces, which are directly influenced by our learned simulation parameter function $f_{\bm{\theta}}$, $\{\bm{q},\dot{\bm{q}}\}$ is the state vector, $\bm{M}$ is the mass matrix, $\bm{J}$ is the Jacobian, and $\bm{C}$ is the Coriolis force. As we will show in the experiments, dynamics discrepancies not included in $\bm{c}$ such as inertial differences, or Young’s modulus of a deformable surface, can also be ``absorbed'' by our state-dependent $f$ (hence state-dependent $\bm{\bm{\tau}}$ and $\bm{p}$). Learning an $f$ with non-vanishing variance $\bm{\sigma}_t$ makes our hybrid simulator stochastic and improves the robustness of the refined policy.
%%SL.10.30: If we have some space, I would recommend expanding this section. It's quite important, and currently it's a bit terse and hard to understand.

% \subsection{Adaptive Alive Bonus}

% One problem we encounter with the Adversarial RL formulation is that it implicitly assumes all expert trajectories have similar or infinite lengths. This problem would not manifest itself clearly when, for example, we use Adversarial RL to solve for an optimal locomotion policy where the given expert trajectories can successfully run indefinitely. However in our case, the collected target domain trajectories are from a non-deterministic suboptimal policy and have very different lengths before termination (in case of locomotion, robot falling over). Therefore the dynamics function $f$ should be optimized so that it is able to generate trajectories with different lengths given different action sequences, with the length distribution matching the collected ones. The cause of this issue lies in the fact that in episodic RL, all transition tuples after trajectory termination are implicitly assigned the reward value of zero. Depending on whether the average reward $r=\log D - \log (1-D)$ at current iteration tends to be positive or negative, it implicitly encourages the learned model to always generate longer trajectories at the cost of fidelity, or to always “commit suicide” quickly and fail the training. 

\section{Experiment Setup}

We set up our experiments to evaluate if our method can (1) improve domain adaptation for robots with different morphologies (2) handle dynamics discrepancies that are important in sim-to-real transfer (3) handle dynamics discrepancies not  intuitively mapped to the list of parameters that our model identifies but can be absorbed by the state-dependent nature of our model. We also lay out the implementation details of our method.
% We present the setup of the simulated robots, their control policies, and the environments used for evaluation here. We also lay out the implementation details of our method. \todo{Write a few sentences that gives more information, such as the purpose of the experiments.}
%%SL.10.30: It's generally good not to start a section with a subsection
\subsection{Robots and Control Policies}

We evaluate the performance of our framework both on standard Gym \cite{1606.01540} benchmarks and on robots of real-world interest. We utilize two simulated robots in our experiments: a 4-link, 6-DoF 2D hopper and the 18-DoF Unitree Laikago quadruped \cite{laikago2018}. The policy for each robot is a 3-layer feed-forward neural network, where the input is robot observation, and the output is the motor torque. The observation space of the hopper is the full robot state to match implementations in OpenAI Gym \cite{1606.01540}. The observation space of the Laikago robot includes sensor measurements and state estimations we can collect from the physical robot, including: root global orientation, root height, root linear velocities and the 12-DoF joint angles. A 10\% white noise is added to each observation, and a 5\% white noise is added to each torque command. The control frequency for both robots is set to 50Hz. 

In most of the experiments except in the multi-task setting (Sec.\ref{sec:cross-task}), we choose a reward function that encourages the robot to move forward with smooth joint motion and low energy:
\begin{equation*}
    r(\bm{s},\bm{a}) = w_c + w_v v_x - w_a \norm{\bm{a}}^2 - w_j\norm{\bm{j}}_0 - w_s\norm{\Ddot{\bm{q}}}, 
\end{equation*}
where $v_x$ is the root forward velocity, $\norm{\bm{j}}_0$ is number of joints at joint limit, and $w_c,w_v, w_a, w_j, w_s$ are the weights. These weights are tuned differently for Hopper and Laikago robots to learn a behavior policy $\pi_B$ that is optimal in the source domain (suboptimal in target domains), and are kept unchanged for policy refinement and baseline methods. 

\subsection{Target Domain Setup}

%Hard enough
We use PyBullet \cite{coumans2017PyBullet} as the simulation platform in this work. The source environment contains the robot and a rigid ground plane in PyBullet. We set up the following target environments for domain transfer:

\textbf{Deform:} Contacts are difficult to model but play a key role in locomotion. For this reason, we first test our method for adaptation to different contact dynamics. We set up a large mattress-like deformable cube in PyBullet, replacing the hard floor for the robots to locomote on. This gap is also chosen to test the expressiveness of our method, as deformable bodies are simulated in PyBullet independently with FEM methods, hence their physical properties are controlled by a set of parameters that are not intuitively mapped to the list of contact parameters that our model identifies.

\textbf{Power:} Actuator dynamics is one of the major causes of the sim-to-real gap \cite{tan2018simtoreal, hwangbo2019learning}. Therefore, for the Laikago, we set up a target environment where the actual torques are the commanded ones subtracting $c \dot{\bm{q}}$, mimicking unmodeled motor friction and Back-EMF torques \cite{tan2018simtoreal}. For the Hopper, as this discrepancy is not hard enough to fail the behavior policy, we halve its ankle torque, mimicking a broken motor.

\textbf{Heavy:} It is important that our model can handle domain shifts that are not caused by actuator or contact dynamics, as discussed in Sec. \ref{sec:parameterization}. Here we evaluate if our identified state-action-dependent actuator and contact parameters can absorb changes in inertia. For the Laikago, we increase the masses of its front-left leg links by 3kg and move the Center of Mass (CoM) of the leg down by 10cm; for the Hopper, we increase the masses of its torso and thigh by 3kg in total and move the CoM of both links up by 10cm.

\subsection{Implementation of Algorithm}

We use a policy that is trained to saturation in the source environment as $\pi_B$. During data-collection using $\pi_B$, we add Gaussian noise with standard deviation of 0.25 to each policy action to make the recorded dataset more diverse. In each of the target environments, we collect 200 trajectories. The lengths of these trajectories (time before falling) ranges between one to two seconds. In total, the target environment dataset consists of 10,000 to 20,000 environment steps. 

We model the simulator parameter function $f$ as a 3-layer two-branch feed-forward neural network with $\tanh$ activations, with each branch outputting contact and actuator parameters (mean and log-variance) separately. We use large output range scaling for $f$ without tuning. Differing from policies with partial observations, the input to $f$ is the full state $\bm{s}_t$ concatenated with actions from $\pi_B(\bm{a}_t|\bm{o}_t)$. We model the discriminator as a 3-layer neural network as well. We train $f$ with PPO \cite{schulman2017proximal} until the average discriminator score converges, which takes roughly 200 iterations for Hopper and 1,000 iterations for Laikago. Between each RL iteration, the discriminator is trained for five epochs.
% there is also alive bonus that tends to fluctuate, so let's not say return converges here 
% I am leaning towards not overloading f with a new name (generator) as it already has a formal name (simulation parameter function)

During the policy refinement step, no more data in the target domain is collected and $f$ is kept unchanged. To re-train $\pi_{new}$ under the dynamics of learned $f$, we warm-start $\pi_{new}$ from $\pi_B$ with halved learning rate. As the training environment has changed drastically, we only inherit the policy weights but re-initialize the value function. Consistent with the training curves of the behavior policies, Laikago takes more PPO iterations (500) than the Hopper (250) before learning converges.

\section{Evaluations}

We evaluate our method on the aforementioned robots and target domains with the following baseline methods:
\begin{enumerate}
    \item Fine-tuning from $\pi_B$ (FT): We directly fine-tune $\pi_B$ in the target environments with the 200 trajectory budget in each environment, which is the same data budget as in our method. However, as the policy improves, FT generally produces longer rollouts and ends up using more samples than our method. We ran a grid search for the batch-size to optimize data-efficiency when using PPO to train FT.
    \item Domain randomization (DR): The selected parameters for DR and their ranges are listed in Table \ref{tab:params}. For Laikago, we inherit the selection and ranges from previous works \cite{RoboImitationPeng20, tan2018simtoreal} and add three contact-related parameters to match the output of $f$ and to account for the contact gaps. The same selection is used for Hopper. For parameter ranges not listed in the previous papers, we manually search for the largest ranges that allow for successful policy training. We ran a grid search for the best-performing \emph{learning} parameters.
    \item DR + FT: We fine-tune the Domain Randomization policies in the target environments for 200 trajectories.
    \item System ID with recorded open-loop actions (SysID-o): Using the same 200-trajectory dataset from $\pi_B$, we execute the recorded action sequences from the same initial states in the simulator. Following \cite{chebotar2019closing}, we optimize the \textit{state-independent} means and variances of parameters used in $f$. The same $l_p$-norm fitness function and trajectory Gaussian filtering are used and the parameters are optimized by CMA-ES \cite{hansen2019pycma}. We then re-train from $\pi_B$ under the learned parameter distribution.
    \item System ID with closed-loop actions (SysID-c): This baseline is same as above except that the rollouts in the simulator use closed-loop actions from $\pi_B$.
\end{enumerate}
Additionally, we demonstrate the cross-task generalization of our hybrid simulator: we show that our hybrid simulator learned with data from one behavior policy can be used for learning different motor tasks. Finally, we conduct ablation studies to understand the impact of different design decisions, including data budget, alive bonus, and more task-relevant data. The qualitative results are presented in the accompanying video.

\begin{table}[h]

\caption{Dynamics parameters randomization ranges.}
\label{tab:params}

\centering
\begin{tabular}{lll}
\toprule
\textbf{Parameter} & \textbf{\begin{tabular}[c]{@{}l@{}} DR Range \\ Hopper \end{tabular}} & \textbf{\begin{tabular}[c]{@{}l@{}} DR Range \\ Laikago \end{tabular}} \\ \midrule
Mass Ratio        & {[}0.5, 1.5{]} & {[}0.8, 1.2{]} \\ %\hline
Inertia Ratio      & {[}0.4, 1.8{]} & {[}0.5, 1.5{]} \\ %\hline
Motor Scaling     & {[}0.5, 1.5{]} & {[}0.8, 1.2{]}  \\ %\hline
Motor Friction    & {[}0.2, 3.0{]} & {[}0.2, 2.0{]} \\ %\hline
Latency (ms)      & {[}0, 40{]}  & {[}0, 40{]}  \\ %\hline
Lateral Friction  & {[}0.4, 1.5{]} & {[}0.4, 1.25{]} \\ %\hline
Spinning Friction & {[}0, 0.2{]}   & {[}0, 0.1{]}   \\ %\hline
Restitution       & {[}0, 1.5{]}   & {[}0, 0.5{]}  \\ %\hline
Contact ERP       & {[}15, 1500{]} & {[}100, 1500{]}  \\ \bottomrule
% Mass Ratio        & {[}0.5, 1.5{]} & {[}0.8, 1.2{]} & \\ %\hline
% Inertia Ratio      & {[}0.4, 1.8{]} & {[}0.5, 1.5{]} & \\ %\hline
% Motor Scaling     & {[}0.5, 1.5{]} & {[}0.8, 1.2{]} & {[}-0.5, 1.5{]} \\ %\hline
% Motor Friction    & {[}0.2, 3.0{]} & {[}0.2, 2.0{]} & \\ %\hline
% Latency (ms)      & {[}0, 40{]}  & {[}0, 40{]} &  \\ %\hline
% Lateral Friction  & {[}0.4, 1.5{]} & {[}0.4, 1.25{]} & {[}0, 5{]} \\ %\hline
% Spinning Friction & {[}0, 0.2{]}   & {[}0, 0.1{]} &  {[}0, 5{]}  \\ %\hline
% Restitution       & {[}0, 1.5{]}   & {[}0, 0.5{]}  &  {[}0, 10{]} \\ %\hline
% Contact ERP       & {[}15, 1500{]} & {[}100, 1500{]} & {[}10, 2000{]} \\ \bottomrule
\end{tabular}
\end{table}

\subsection{Performance in the Target Domains}
\label{sec:mean-results}
% \begin{figure}[thpb]
%   \centering
% %   \framebox{\parbox{3in}{We suggest that you use a text box to insert a graphic (which is ideally a 300 dpi TIFF or EPS file, with all fonts embedded) because, in an document, this method is somewhat more stable than directly inserting a picture.
% % }}
%   \includegraphics[scale=0.25]{table1-draft.png}
%   \label{figurelabel}
% \end{figure}

\begin{table*}[h]
\caption{Average task reward in six domain adaptation experiments.}
\label{tab:results}
\begin{center}
\centering
\begin{tabular}{llllll}
\toprule
\textbf{Environment} & \textbf{$\pi_B$} & \textbf{Ours} & \textbf{FT} & \textbf{DR} & \textbf{DR+FT} \\
\midrule
Hopper/Deform & 205 & \textbf{1389 $\pm$ 148} & 622 $\pm$ 78 & 815 $\pm$ 422  & 996 $\pm$ 530 \\
Hopper/Power & 178 &
\textbf{1601 $\pm$ 19} &
411 $\pm$ 75 &
590 $\pm$ 123 &
1237 $\pm$ 466
\\
Hopper/Heavy & 223 &
1658 $\pm$ 71 &
\textbf{1986 $\pm$ 38} &
807 $\pm$ 254 &
1905 $\pm$ 193
\\
Laikago/Deform & 481 &
\textbf{2650 $\pm$ 220} &
2023 $\pm$ 357 &
2206 $\pm$ 204 &
2231 $\pm$ 152
 \\
Laikago/Power & 271 &
\textbf{2484 $\pm$ 106} &
908 $\pm$ 73 &
2408 $\pm$ 153 &
2415 $\pm$ 144
\\
Laikago/Heavy &  308 &
\textbf{2211 $\pm$ 326} &
550 $\pm$ 124 &
1523 $\pm$ 283 &
1865 $\pm$ 183
\\
\bottomrule
\end{tabular}
\end{center}
\end{table*}

% \begin{table}[h]
% \caption{Mean task reward of system id.}
% \label{tab:sysid}
% \begin{center}
% \centering
% \begin{tabular}{lll}
% \toprule
% \textbf{Environment} & \textbf{SysID-o} & \textbf{SysID-c} \\
% \midrule
% Laikago/Power & 189     & 199     \\
% \bottomrule
% \end{tabular}
% \end{center}
% \end{table}

As shown in Table \ref{tab:results}, our method outperforms the baseline methods on five of the six domain adaptation experiments. Each number in Table \ref{tab:results} shows the average task reward in the target environment over 30 rollouts from random initial states. For each method, we report the standard deviation of five trials with different random seeds. 

One advantage of our method over DR is that DR tends to produce policies that trade performance for robustness. For example, in the Laikago/Deform environment, our framework learns a walking gait that is 1.3x faster than the DR policies. As shown in the accompanying video, with the same reward function, DR learns to use conservative, small and frequent gaits, while the proposed method adopts agile galloping gaits with larger strides.

For the System ID baselines, we are unable to identify simulator parameters useful for improving policy performance in the target domain. For example, the task rewards of both variants of System ID are an order of magnitude lower (SysID-o: 189, SysID-c: 199) than our method in the Laikago/Power environment, regardless of hyper-parameters. Our investigation suggests that the System ID trajectories generated from the behavior policy are highly sensitive to initial conditions and physical parameters, possibly because even small errors in contact-rich locomotion environments can make trajectories diverge quickly. Therefore, executing the same open-loop control sequence in simulator and target domains will lead to mismatched observation sequences, and executing the same closed-loop policy will lead to mismatched observation and control sequences. In both cases, such large mismatch invalidates pairwise trajectory-based similarity metrics. Different from $l_p$-norms, our method does not require pairwise trajectory comparison, and instead adopts distributional supervision via GAN loss.

\subsection{Cross-task Generalization}
\label{sec:cross-task}

Our identified hybrid simulator can generalize and be used to learn different tasks, instead of overfitting to the task that the behavior policy solves. We use the same hybrid simulator that is learned for the Deform environment using the data collected by the behavior policy of the Laikago walking forward. But, during the policy refinement, we change the reward function to encourage sideway walk. After the policy refinement, the Laikago is able to walk sideways in the target domain (see accompanying video). The result suggests that the learned hybrid simulator is generalizable, and can be used for learning multiple different motor skills.

% note: makes no sense to compare reward with locomoting forward as gaits are different.

\subsection{Ablation Studies}

We conduct three ablation studies in two domain adaptation tasks to gain more insight into our method. The results below are averaged across three random seeds.

\textbf{Data budget in target domain:}
Table \ref{tab:ablation_trajs} shows the average reward of refined policies in the target environments, when varying the number of target environment trajectories used for learning the simulator. We are glad to find that our method could potentially use a even smaller amount of data in the target domain (e.g. with 50 trajectories) without noticeable performance deterioration.

\begin{table}[h]
\caption{Average task reward in target domains with varying numbers of target trajectories.}
\label{tab:ablation_trajs}
\centering
\begin{tabular}{lllll}
\toprule
\textbf{Num Trajs}    & \textbf{200}  & \textbf{100}  & \textbf{50}   & \textbf{25}   \\ \midrule
Hopper/Heavy & 1658 & 1554 & 1686 & 1551 \\ 
Laika/Deform & 2650 & 2809 & 2561 & 2272 \\ \bottomrule
\end{tabular}
\end{table}

\textbf{Adaptive alive bonus:} Table \ref{tab:ablation_alive} shows the average reward of refined policies in the target environments, with or without adaptive alive bonus $b_i$ applied during the learning of the hybrid simulator. In the Hopper/Heavy environment, adaptive alive bonus helps to improve the simulation fidelity and thus increases the adapted policy performance.

\begin{table}[h]
\caption{Average task reward in target domains with variation in alive bonus.}
\label{tab:ablation_alive}

\centering
\begin{tabular}{lll}
\toprule
             & \textbf{With $b_i$} & \textbf{Without $b_i$} \\ \midrule
Hopper/Heavy & 1658       & 1186          \\ 
Laika/Deform & 2650       & 2569          \\ \bottomrule
\end{tabular}
\end{table}

\textbf{More iterations of Algorithm 1:} One possible way to improve our framework is to run the whole Algorithm 1 for another iteration by using the refined policy as the new behavior policy in the second iteration. We find that this does not clearly improve the performance of the adapted policies. Since our hybrid simulator still uses the analytical simulation model as a strong prior and we have shown that it is already robust to the state-action distribution shift, collecting more task-specific data does not necessarily help in our case.
% our method is not sensitive to the task-relevance of the collected data. After all, we have shown in Table \ref{} that our algorithm can just use a poorly performed behavior policy and improve task performance by several times.

% \begin{table}[h]
% \caption{Mean task reward in evaluation environment of SimGAN with varied adaptation iterations.}
% \label{tab:ablation_iteration}
% \centering
% \begin{tabular}{lll}
% \toprule
%               & \textbf{1 Iteration} & \textbf{2 Iterations} \\ \midrule
% Hopper/Deform & 1389        & 1375         \\ 
% Laika/Heavy   & 2211        & 2108         \\ \bottomrule
% \end{tabular}
% \end{table}

\section{Discussions}
\label{sec:discussions}
In this paper, we tackle the challenge of transferring policies to a new domain with different dynamics, by identifying a hybrid simulator that better matches target domain trajectories. We propose to formulate simulation identification as an adversarial reinforcement learning problem. We use a learned GAN loss in place of the standard mean-squared-errors to measure the discrepancies between distributions of transition tuples, and augment a classical physics simulator with a learned state-action-dependent stochastic simulator parameter function to improve its expressiveness to simulate unmodeled phenomena. Our method shows promising results on six difficult dynamical domain adaptation tasks, outperforming strong baseline methods.

We find that a good parameterization of the hybrid simulator plays a critical role in our system. Before we narrowed down to use contact parameters and motor scaling as the parameterization, we experimented with other alternatives. For example, we tried directly learning all contact forces, similar to \cite{jeong2019modelling}. But, it performed much worse because there is no constraint to prevent energy injection into the physical system. Consequently, the locomotion policy learned to exploit these fictitious forces in simulation which did not exist in the target domain. An interesting future direction would be to experiment with different parameterizations and add physical constraints to the learned components of the hybrid simulator, so that fundamental physical laws, such as energy conservation, are always respected.

Our current implementation has several limitations. First, the behavior policy cannot perform too poorly in the target domain. It needs to collect long enough trajectories (longer than one second at least) before the robot falls so that the trajectories are informative for learning the discriminator and the simulation parameter function. Second, while we keep the hyperparameters the same across all three target environments, they are currently tuned differently for the two robots, which might increase manual effort. Third, while the results for sim-to-sim transfer are promising, due to COVID-19, we are not able to test our method on real robots. In the future, we plan to address these limitations, and test our method for transferring locomotion policies from simulation to a real Laikago robot.

% \begin{eqnarray*}
% (\bm{o}_R, \bm{a}_R, \bm{o}_R^+) \\
% \\
% (\bm{o}, \bm{a}, \bm{o}^+) \\
% \\
% r = \log (D / (1-D))
% \end{eqnarray*}

% \section{Conclusions}
% WIP

\newpage
% \addtolength{\textheight}{-6cm}   % This command serves to balance the column lengths

%%%%%%%%%%%%%%%%%%%%%%%%%%%%%%%%%%%%%%%%%%%%%%%%%%%%%%%%%%%%%%%%%%%%%%%%%%%%%%%%

%%%%%%%%%%%%%%%%%%%%%%%%%%%%%%%%%%%%%%%%%%%%%%%%%%%%%%%%%%%%%%%%%%%%%%%%%%%%%%%%

%%%%%%%%%%%%%%%%%%%%%%%%%%%%%%%%%%%%%%%%%%%%%%%%%%%%%%%%%%%%%%%%%%%%%%%%%%%%%%%%

% \section*{Acknowledgement}

% The preferred spelling of the word ÒacknowledgmentÓ in America is without an ÒeÓ after the ÒgÓ. Avoid the stilted expression, ÒOne of us (R. B. G.) thanks . . .Ó  Instead, try ÒR. B. G. thanksÓ. Put sponsor acknowledgments in the unnumbered footnote on the first page.

%%%%%%%%%%%%%%%%%%%%%%%%%%%%%%%%%%%%%%%%%%%%%%%%%%%%%%%%%%%%%%%%%%%%%%%%%%%%%%%%

% References are important to the reader; therefore, each citation must be complete and correct. If at all possible, references should be commonly available publications. \cite{coumans2017PyBullet}

% \begin{thebibliography}{99}
% \bibitem{c20} J. P. Wilkinson, Nonlinear resonant circuit devices (Patent style),Ó U.S. Patent 3 624 12, July 16, 1990. 
% \end{thebibliography}

\bibliographystyle{IEEEtran} % use IEEEtran.bst style
\bibliography{IEEEabrv,root}

\end{document}